\documentclass[lettersize,journal]{IEEEtran}

\usepackage{amsmath,amsfonts}
\usepackage{array}
\usepackage{textcomp}
\usepackage{stfloats}
\usepackage{url}
\usepackage{verbatim}
\usepackage{graphicx}
\def\BibTeX{{\rm B\kern-.05em{\sc i\kern-.025em b}\kern-.08em
    T\kern-.1667em\lower.7ex\hbox{E}\kern-.125emX}}
\usepackage{balance}

\usepackage{float}
\usepackage{listings}
\usepackage{tcolorbox}
\usepackage{fancyvrb}  
\usepackage{xcolor}
\usepackage{pifont}   
\usepackage{amsmath}
\usepackage{algorithm}
\usepackage{algpseudocode}
\usepackage{amssymb}
\usepackage{makecell}
\usepackage{xspace}
\usepackage{multirow}
\usepackage{booktabs}
\usepackage{subcaption}
\usepackage{hyperref}
\usepackage{cleveref}
\newcolumntype{C}[1]{>{\centering\arraybackslash}m{#1}}
\usepackage{graphicx}
\usepackage{array}
\usepackage{longtable}

\newcommand{\existsmark}{\ensuremath{\bullet}}
\newcommand{\missingmark}{\ensuremath{\circ}}
\newcommand{\cmark}{\textcolor{green!60!black}{\checkmark}} 
\newcommand{\xmark}{\textcolor{red}{\ding{55}}}
\DeclareMathOperator*{\argmax}{arg\,max}

\newcommand{\shorttitle}{\textsc{PragmaBot}\xspace}
\usepackage{soul}
\usepackage{siunitx}

\begin{document}

\title{A Pragmatist Robot: Learning to Plan Tasks by \\Experiencing the Real World}




\author{Kaixian Qu, Guowei Lan, René Zurbrügg, Changan Chen, \\
Christopher E. Mower, Haitham Bou-Ammar, and Marco Hutter
\thanks{K. Qu, G. Lan, R. Zurbrügg, C. Chen, and M. Hutter are with the Robotic Systems Lab, ETH Zürich, Switzerland. R. Zurbrügg is also with the ETH AI Center, ETH Zürich, Switzerland. C. E. Mower and H. Bou-Ammar are with Huawei Noah’s Ark Lab, London, UK. H. Bou-Ammar is also with the UCL Centre for AI, London, UK. Corresponding author: K. Qu (e-mail: kaixqu@ethz.ch).}
\thanks{This research was supported by the Swiss National Science Foundation through the National Centre of Competence in Digital Fabrication (NCCR dfab), by Huawei Tech R\&D (UK) through a research funding agreement, by an ETH RobotX research grant funded through the ETH Zürich Foundation, and partially by the ETH AI Center. This work was also conducted as part of ANYmal Research, a community to advance legged robotics. We would also like to thank Cesar Cadena for his support and helpful discussions.}
\vspace{-1em}
}

\markboth{Journal of \LaTeX\ Class Files,~Vol.~18, No.~9, September~2020}%
{A Pragmatist Robot: Learning to Plan Tasks by Experiencing the Real World}

\maketitle
\thispagestyle{plain}
\pagestyle{plain}

\begin{abstract}
Large language models (LLMs) have emerged as the dominant paradigm for robotic task planning using natural language instructions. However, trained on general internet data, LLMs are not inherently aligned with the embodiment, skill sets, and limitations of real-world robotic systems. Inspired by the emerging paradigm of verbal reinforcement learning---where LLM agents improve through self-reflection and few-shot learning without parameter updates---we introduce \shorttitle, a framework that enables robots to learn task planning through real-world experience. \shorttitle employs a vision-language model (VLM) as the robot's ``brain'' and ``eye'', allowing it to visually evaluate action outcomes and self-reflect on failures. These reflections are stored in a short-term memory (STM), enabling the robot to quickly adapt its behavior during ongoing tasks. Upon task completion, the robot summarizes the lessons learned into its long-term memory (LTM). When facing new tasks, it can leverage retrieval-augmented generation (RAG) to plan more grounded action sequences by drawing on relevant past experiences and knowledge. Experiments on four challenging robotic tasks show that STM-based self-reflection increases task success rates from $35\%$ to $84\%$, with emergent intelligent object interactions. In 12 real-world scenarios (including eight previously unseen tasks), the robot effectively learns from the LTM and improves single-trial success rates from $22\%$ to $80\%$, with RAG outperforming naive prompting. These results highlight the effectiveness and generalizability of \shorttitle. Project webpage: \texttt{https://pragmabot.github.io/}
\end{abstract}






\section{Introduction}

\IEEEPARstart{R}{ecently}, large language models (LLMs) have demonstrated near-human performance across a range of reasoning tasks, showcasing emergent capabilities in diverse domains such as coding and law~\cite{ chen2021evaluating, wei2022chain, ahn2024large, fei2023lawbench}. 
These broad competencies have enabled LLMs to move beyond traditional language tasks and sparked interest in the field of robotics.
In particular, they are now widely used in task planning, where LLMs interpret natural language instructions and generate feasible action plans with common-sense reasoning~\cite{liang2023code, ahn2022can, mower2024ros, huang2023inner, huang2023voxposer}. However, applying LLMs to robotics remains challenging, as they are designed exclusively for text processing, while robots must operate based on continuous, high-dimensional sensor streams.

To address the limitations of text-only input, research has increasingly shifted toward multimodal approaches, especially vision-language models (VLMs) that jointly process visual and textual data. Recent VLMs~\cite{achiam2023gpt, bubeck2023sparks, team2024gemini} exhibit strong multimodal reasoning and high-resolution visual processing. Building on these capabilities, recent work has leveraged VLMs to enable robots to reason about visual inputs and operate in closed-loop, autonomous settings~\cite{brohan2023rt, zhi2024closed, mei2024replanvlm, hu2023look}.

\begin{figure}[t]
    \centering
    \includegraphics[width=\linewidth]{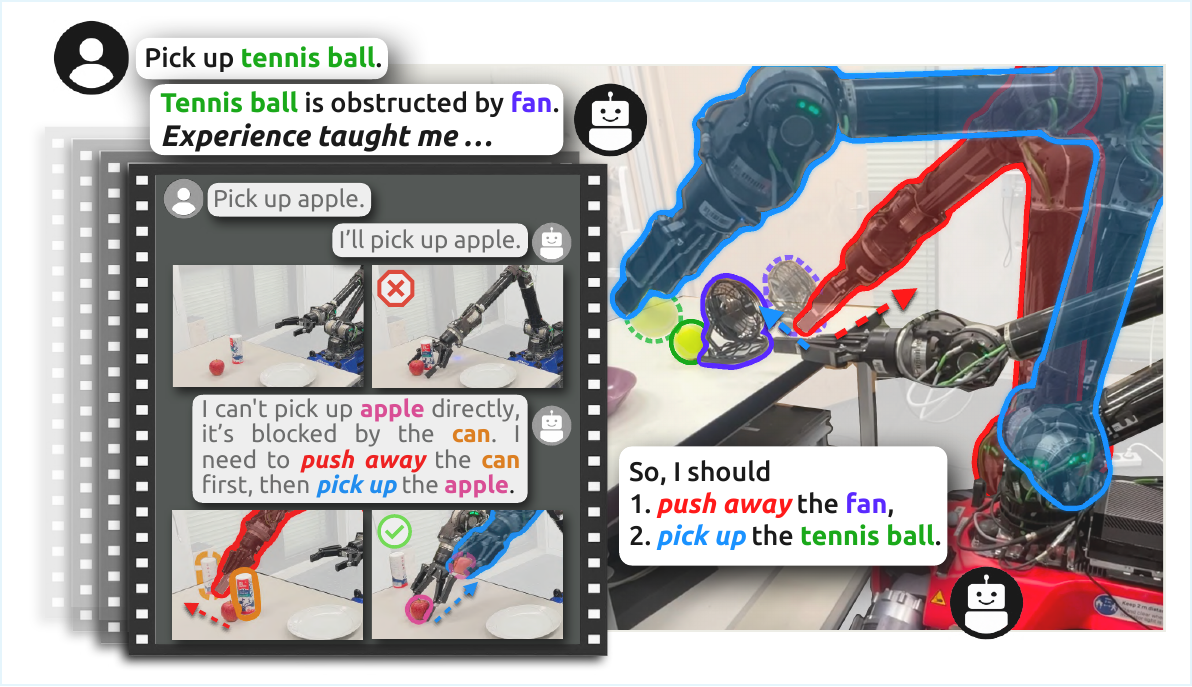}
    \caption{
    \textbf{Robot completes a new task guided by a long-term memory of self-reflective experiences.}
    When executing a novel task, the robot maintains a short-term memory that helps it reflect and learn how to complete the task (illustrated in the grey clip).
    The experience is then stored as long-term memory and retrieved to guide the VLM’s task planning whenever a similar scenario is encountered (illustrated in the main figure).}
    \label{fig:teaser}     
    \vspace{-1.5em}
\end{figure}

Yet, transferring these internet-trained models to physical robots remains challenging. While VLMs excel at abstract reasoning and visual understanding, they are not inherently aligned with the embodiment, skill sets, and limitations of real-world robotic systems.
For example, in~\Cref{fig:teaser}, a VLM may confidently instruct a grasp on a partially occluded tennis ball, consistent with human intuition, yet the robot may fail due to limited manipulation capabilities under partial observability.
This raises a critical question:
How can the robot align the VLM with its own capabilities to achieve better task planning performance?

This paper introduces \shorttitle, a framework that enables robots to learn task planning through real-world experience. 
Our method is inspired by the principles of verbal reinforcement learning~\cite{shinn2023reflexion}, where the LLM improves its performance via in-context learning on past self-reflective experience, without incurring the expense of fine-tuning the LLM through weight updates.
\shorttitle utilizes a VLM in three different ways: 
(i) to plan actions for execution, 
(ii) to verify that the action was completed successfully, and
(iii) to summarize experiences.
It maintains a short-term memory (STM) that keeps track of executed actions and associated feedback signals, along with a long-term memory (LTM) that stores lessons learned from past successful task executions.
\shorttitle uses the STM to perform self-reflection, allowing the robot to adapt its behavior toward achieving the task goal. Upon successful task completion, the VLM summarizes the STM experience and stores it in the LTM. 
When presented with a similar task in the future, the robot employs retrieval-augmented generation (RAG) to retrieve relevant knowledge from the LTM for planning actions while accounting for its own capabilities and limitations.
Additionally, \shorttitle enhances the spatial understanding of VLMs through an on-demand image annotation module.

This paper presents the following key contributions:
\begin{itemize}
    \item We introduce \shorttitle, which adapts verbal reinforcement learning to real-world robotic task planning by integrating VLMs with visual feedback. We demonstrate that the combination of short-term and long-term memory enables efficient learning of task planning tailored to the robot's embodiment and capabilities.
    \item We introduce retrieval-augmented generation (RAG) into the verbal reinforcement learning framework, allowing the robot to selectively leverage task-relevant, self-reflective experiences. Extensive evaluations show that RAG improves planning accuracy significantly compared to naive prompting.
    \item We design an on-demand image annotation module that enhances the VLM’s spatial reasoning across diverse skills. We demonstrate that this module leads to more accurate action execution in complex, real-world environments.
\end{itemize}

Extensive real-world experiments show that \shorttitle significantly outperforms state-of-the-art methods in task success rates and generalizes well to similar but previously unseen tasks. To support future research, our code is available on the project webpage\footnote{\texttt{https://pragmabot.github.io/}}.

\section{Related Work}
\label{sec:related_work}

Recent advances in robotic task planning have explored methods enabling robots to understand their own capabilities and limitations. SayCan~\cite{ahn2022can} grounds language understanding in robotic affordances by combining LLM outputs with a trained visual affordance network, which requires extensive training resources. Alternative frameworks focus on building robotic memory through dense human feedback and corrections to improve decision-making~\cite{zha2024distilling, barmann2024incremental}. Building on this direction, BUMBLE~\cite{shah2024bumble} integrates short-term memory for online replanning with long-term memory of human-annotated failure cases to guide VLMs in avoiding past errors. However, these methods typically rely heavily on dense human supervision for feedback and correction.

More recent work explores autonomous failure recovery through self-reflection mechanisms. Systems such as~\cite{liu2023reflect, wang2024can} enable LLM/VLM-powered robots to analyze their own failures and adapt subsequent actions to accelerate task completion. Building on this idea, Reflexion~\cite{shinn2023reflexion} goes further by not only allowing the LLM agent to reflect on its failures but also storing these self-reflective experiences in long-term memory to inform future behavior. This approach, known as verbal reinforcement learning, improves agent performance not by updating model weights, but by incorporating additional contextual information for reasoning. While Reflexion eliminates the need for extensive training data, it has so far been evaluated primarily in simulated environments that do not account for real-world embodiment or the complexities of physical interaction. Moreover, observations must be supplied either as ground truth from the simulator or via external, hand-engineered scene descriptors.

\begin{table}[t]
\centering
\footnotesize
\caption{\textbf{Comparison with baselines.} \shorttitle is the only method that achieves self-reflection, learning by experiencing, interactive replanning, and creative tool use (highlighted in bold). System components are listed in regular font (\existsmark\, indicates presence).}
\begin{tabular}{@{}l
                >{\centering\arraybackslash}p{0.20cm}
                >{\centering\arraybackslash}p{0.20cm}
                >{\centering\arraybackslash}p{0.20cm}
                >{\centering\arraybackslash}p{0.20cm}
                >{\centering\arraybackslash}p{0.20cm}
                >{\centering\arraybackslash}p{0.20cm}
                >{\centering\arraybackslash}p{0.20cm}
                >{\centering\arraybackslash}p{1.80cm}
                @{}}\toprule
&  \rotatebox{30}{\makecell{\textbf{Self-reflection}}}
&  \rotatebox{30}{\makecell{\textbf{Learning by exp.}}} 
&  \rotatebox{30}{\makecell{\textbf{Interactive replan}}} 
&  \rotatebox{30}{\makecell{\textbf{Creative tool use}}} 
&  \rotatebox{30}{\makecell{Short-term memory}} 
&  \rotatebox{30}{\makecell{Long-term memory}}
&  \rotatebox{30}{\makecell{Unified visual feedback}} 
\\
\midrule
CaP~\cite{liang2023code}            & \xmark & \xmark & \xmark & \xmark & \missingmark & \missingmark & \missingmark & \\
SayCan~\cite{ahn2022can}            & \xmark & \xmark & \xmark & \xmark & \existsmark & \missingmark & \missingmark & \\
Inner Mono.~\cite{huang2023inner}   & \xmark & \xmark & \xmark & \xmark & \existsmark & \missingmark & \missingmark & \\
RoboTool~\cite{xu2023creative}      & \xmark & \xmark & \xmark & \cmark & \missingmark & \missingmark & \missingmark & \\
DROC~\cite{zha2024distilling}       & \xmark & \cmark & \xmark & \xmark & \existsmark & \existsmark & \missingmark &  \\
REFLECT~\cite{liu2023reflect}       & \cmark & \xmark & \cmark & \xmark & \existsmark & \missingmark & \missingmark &  \\
COME~\cite{zhi2024closed}           & \cmark & \xmark & \xmark & \xmark & \existsmark & \missingmark & \existsmark &  \\
ReplanVLM~\cite{mei2024replanvlm}   & \cmark & \xmark & \cmark & \xmark & \existsmark & \missingmark & \existsmark &  \\
BUMBLE~\cite{shah2024bumble}        & \cmark & \xmark & \cmark & \xmark & \existsmark & \existsmark & \missingmark &  \\
\hline
\shorttitle                          & \cmark & \cmark & \cmark & \cmark & \existsmark & \existsmark & \existsmark & \\
\bottomrule
\vspace{-2em}
\label{tab:expteach-comparison}
\end{tabular}
\end{table}

\shorttitle investigates how physical robots can visually evaluate the outcomes of their actions in challenging tasks, self-reflect on failures, and construct long-term memory to guide future planning---all through real-world experience. It enables the system to learn without relying on explicit human annotation or dense human feedback. Additionally, we incorporate Retrieval-Augmented Generation (RAG) to retrieve relevant past experiences from memory. While RAP~\cite{kagaya2024rap} proposes using RAG with a memory built from diverse experiences for simulated LLM agents, it does not consider self-reflection with VLMs. \Cref{tab:expteach-comparison} compares \shorttitle with prior LLM/VLM task planners. ``Learning by experiencing'' implies that the robot learns from failure, adapts, and memorizes the solution for future planning (with or without human correction). ``Interactive replanning'' refers to interacting with non-target objects after failure to aid task completion. ``Creative tool use'' involves autonomously using unmentioned objects as tools. ``Unified visual feedback'' refers to the robot using the same VLM for both planning and verification, thereby aligning more closely with the concept of self-reflection. 

Beyond task planning, VLMs can also be used to determine precise action parameters through mask-based techniques~\cite{yang2023set, nasiriany2024pivot}. Recent approaches combine object-centric annotations with grasping tools to support challenging semantic manipulation tasks~\cite{fangandliu2024moka, tziafas2024towards, qian2024thinkgrasp}. However, these methods are typically limited to planar tabletop settings or grasping-only scenarios. In contrast, \shorttitle introduces an on-demand image annotation tool that enables grounded actions in 3D space, supporting a broader range of skills beyond 6-DoF grasping.


\section{Problem Formulation}
Consider a robot system equipped with $K$ predefined parameterized skills $\{\pi_k\}_{k=1}^K$, where each $\pi_k$ is a low-level policy mapping observations (e.g., camera inputs) to actionable commands (e.g., joint motions), operating until certain termination conditions are met. These skills are assumed to be provided in advance, either learned through imitation or reinforcement learning, or implemented as optimal controllers such as model-predictive control.

\begin{figure*}[!t]
    \centering
    \begin{minipage}[t]{0.66\textwidth}
        \vspace{0pt} 
        \centering
        \includegraphics[width=\textwidth]{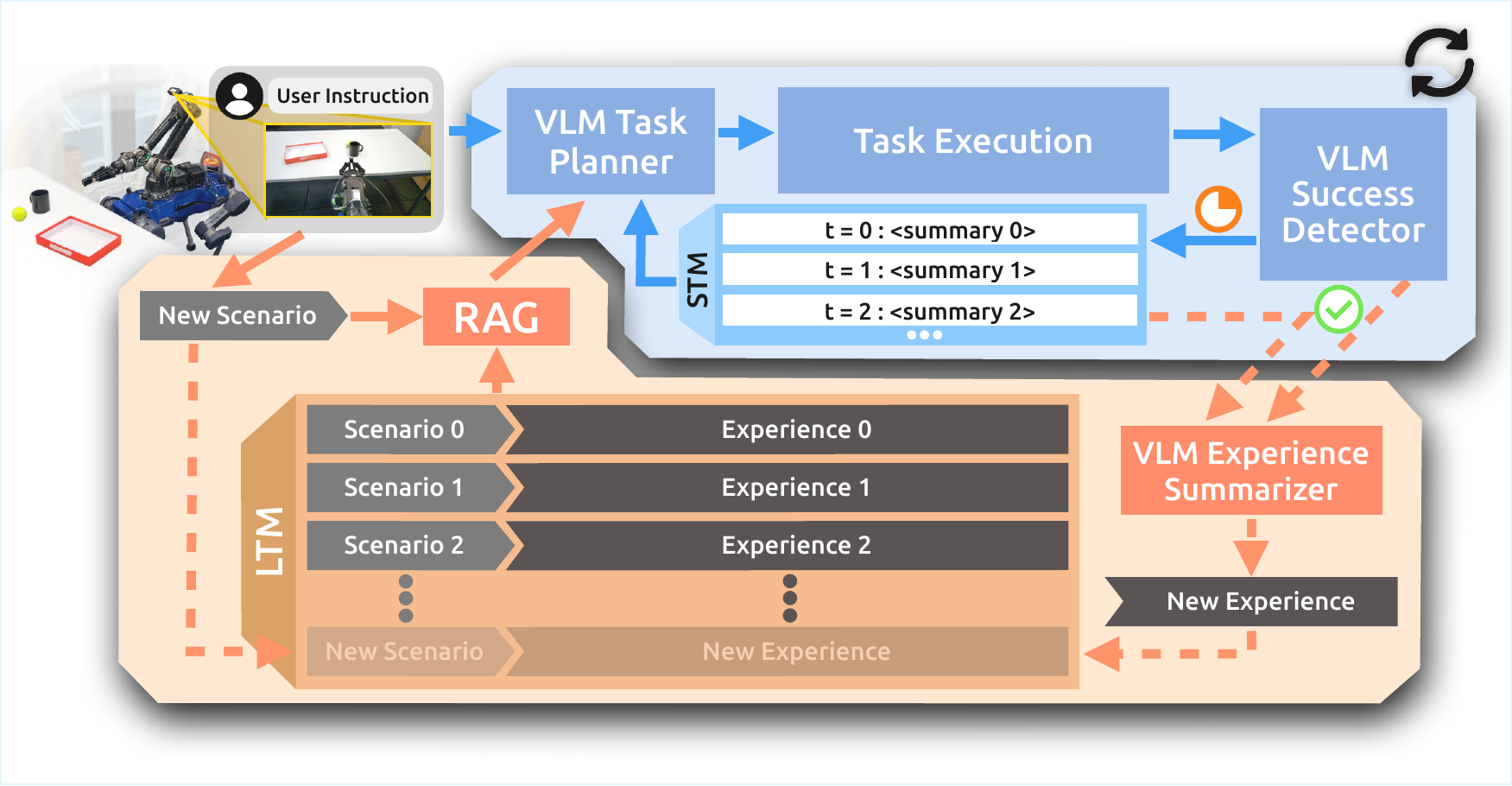}
    \end{minipage}
    \hfill
    \begin{minipage}[t]{0.32\textwidth}
        \vspace{0pt} 
        \begin{algorithm}[H]
            \caption{\shorttitle}
            \label{alg:expteach}
            \footnotesize 
            \setlength{\textfloatsep}{0pt}
            \setlength{\belowcaptionskip}{-5pt}
            \textbf{Given:} Instruction $\mathbf{I}$, initial observation $\mathbf{o}_0$ \\
            \textbf{Internal:} Long-term memory $\mathbf{M}$
            \begin{algorithmic}[1]
                \State $\mathbf{K}' \gets (\mathbf{I}, \mathcal{D}(\mathbf{o}_0))$
                \State $\{(\mathbf{K}_i, \mathbf{E}_i)\}_{i=0}^{k-1} \gets$ 
                \Statex $ \quad \operatorname*{arg\,top\,k}_{(\mathbf{K}, \mathbf{E}) \in \mathbf{M}} \left[ \frac{\mathcal{E}(\mathbf{K})^T \mathcal{E}(\mathbf{K}')}{\|\mathcal{E}(\mathbf{K})\| \|\mathcal{E}(\mathbf{K}')\|} \right]$
                \State $t \gets 0$, $\mathbf{m} \gets \emptyset$
                \Repeat
                    \State $\mathbf{a}_t \gets \mathcal{P}\left(\mathbf{I}, \mathbf{o}_t, \mathbf{m}, \{(\mathbf{K}_i, \mathbf{E}_i)\}_{i=0}^{k-1}\right)$
                    \State Execute $\mathbf{a}_t$ and receive $\mathbf{o}_{t+1}$
                    \State $\mathbf{r}_{t+1} \gets \mathcal{R}(\mathbf{o}_{t}, \mathbf{a}_t, \mathbf{o}_{t+1})$
                    \State $\mathbf{m} \gets \mathbf{m} \cup (\mathbf{a}_t, \mathbf{r}_{t+1})$
                    \State $t \gets t + 1$
                \Until{$\mathbf{r}_t.\text{completed}$}
                \State $\mathbf{M} \gets \mathbf{M} \cup \{(\mathbf{K}', \mathcal{S}(\mathbf{m}))\}$
            \end{algorithmic}
        \end{algorithm}
    \end{minipage}
    \caption{
    \textbf{\shorttitle pipeline (illustration: \textit{left}, algorithm: \textit{right}).}
       At the start of each task, the system takes the user instruction $\mathbf{I}$ and egocentric observation $\mathbf{o}_{0}$, which the VLM combines into a scenario (key). RAG retrieves relevant experiences from long-term memory $\mathbf{M}$ and, together with the instruction and observation, feeds them into the VLM task planner $\mathcal{P}$ to generate the next action $\mathbf{a}$. After execution, success is checked by the VLM success detector $\mathcal{R}$. If the task is not completed, the action $\mathbf{a}$ and its feedback $\mathbf{r}$ are accumulated into short-term memory $\mathbf{m}$ and fed back into planning. Once the task is completed, the short-term memory $\mathbf{m}$ is summarized and stored in long-term memory $\mathbf{M}$ for future use.
    }
    \label{fig:pipeline_overall}
    \vspace{-10pt}
\end{figure*}

We focus on task planning---specifically, how the robot can effectively sequence these skills to accomplish a task described by a human in natural language. To address this, we introduce a higher-level skill selection policy $\Pi$, implemented using foundation models such as VLMs. At each time step, the selected skill is sampled according to
\begin{equation}\label{eq:skill-selection-policy}
    \pi_k \sim \Pi_{\theta}(\mathbf{I}, \mathbf{o}_t, \mathbf{c}_t),
\end{equation}
where $\mathbf{I}$ denotes the natural language instruction, $\mathbf{o}_t$ is the current environmental observation (e.g., an image from the robot’s camera), $\mathbf{c}_t$ represents additional contextual information, and $\theta$ denotes the parameters of the VLM. 
We aim to investigate how to effectively adapt and improve the skill selection policy~\eqref{eq:skill-selection-policy} without requiring extensive training resources or dense human supervision.

\section{Method}

\shorttitle enables the robot to learn to plan tasks through interaction with the real world, following a paradigm akin to reinforcement learning. This is achieved through several core components: a success detector, a memory mechanism, and a memory retrieval strategy. Each component plays a distinct role in the learning pipeline, as detailed below. An overview of the full system is illustrated in \Cref{fig:pipeline_overall}.

\subsection{Task Planner}
\label{sec:task}

The task planner $\mathcal{P}$ leverages a VLM to interpret the user instruction $\mathbf{I}$ and, conditioned on the current RGB observation $\mathbf{o}_t$, outputs the next action $\mathbf{a}_t$---which consists of a selected skill $\pi_k$ along with its associated parameters. This behavior is governed by an underlying policy $\Pi_{\theta}$, which samples actions using the VLM’s visual-language understanding and common-sense reasoning capabilities encoded in its parameters $\theta$. However, this initial policy is insufficient on its own---it lacks awareness of the robot’s capabilities and limitations. To bridge this gap, $\mathcal{P}$ must be improved through real-world interaction, adapting its decisions to better align with the robot’s embodiment.

\subsection{Success Detector}
\label{sec:evaluator}

The success detector $\mathcal{R}$, also implemented as a VLM---the same model used by $\mathcal{P}$---provides a feedback signal $\mathbf{r}$ that assesses the outcome of an executed action $\mathbf{a}$ by comparing the observations before and after the action:
\begin{equation}
    \mathbf{r}_{t+1} \gets \mathcal{R}(\mathbf{o}_{t}, \mathbf{a}_t, \mathbf{o}_{t+1})
\end{equation}
The resulting feedback signal provides two binary indicators: whether the action was successful and whether the overall task is complete. Additionally, it includes a semantic description of the scene changes induced by the action, which can be leveraged by the task planner to determine subsequent actions.

\subsection{Short-Term Memory and Online Adaptation}
\label{sec:stm}

\shorttitle keeps a short-term memory (STM) module, denoted $\mathbf{m}$, which functions as a dynamic log that interacts iteratively with $\mathcal{P}$ and $\mathcal{R}$. It stores a sequence of past interactions up to time $t$, where each entry consists of an executed action $\mathbf{a}_\tau$ and the corresponding feedback signal $\mathbf{r}_{\tau+1}$ generated by $\mathcal{R}$:
\begin{equation}
    \mathbf{m} = \{(\mathbf{a}_\tau, \mathbf{r}_{\tau+1})\}_{\tau=0}^{t-1}.
\end{equation}
Upon action failure, $\mathcal{P}$ performs self-reflection, identifying potential causes and propose better ways to complete the task. This is part of the chain-of-thought before outputting the final action~\cite{wei2022chain}. This process echoes gradient computation in RL, but instead of updating network weights, the planner uses this ``linguistic gradient''---expressed in natural language---to improve its behavior. For example, if the robot fails to pick up an apple partially occluded by a can, the linguistic gradient might read: ``Previous attempts to pick up the apple directly have failed. The apple is next to a cylindrical container, which might be causing interference. To create more space and ensure a successful grasp, I will push the can to the right, away from the apple. This should allow for a clearer path to pick up the apple.'' In addition, we observe that the robot is capable of learning to use tools for manipulating small objects after an initial action failure. This STM serves as a crucial component for online adaptation, effectively substituting weight updates with in-context learning guided by linguistic feedback.

\subsection{Long-Term Memory and Experience Summarization}
\label{sec:ltm}

The STM is episodic and resets upon task completion, thus it does not retain knowledge across tasks.
To enable cumulative learning and transfer of experience over time, we introduce a persistent long-term memory (LTM), denoted $\mathbf{M}$. Upon successful task completion, the robot summarizes its STM into LTM using a VLM-based experience summarizer $\mathcal{S}$. This summary is stored as a key-value pair $(\mathbf{K}, \mathbf{E})$, where the key $\mathbf{K}$ is a scenario description and the value $\mathbf{E} = \mathcal{S}(\mathbf{m})$ is the corresponding summarized experience:
\begin{equation}
 \mathbf{M} \gets \mathbf{M} \cup \{(\mathbf{K}, \mathbf{E})\}.
\end{equation}
The key $\mathbf{K}$ combines two components: (1) the user instruction $\mathbf{I}$ and (2) a natural language description of the initial scene, generated by a VLM scene describer $\mathcal{D}$ from the first RGB observation $\mathbf{o}_0$ (e.g., ``The apple is on the right side of the table, next to a salt container. The plate is on the left side, with ample space around it.''). This combined context provides a rich, semantically meaningful index for future retrieval. To enable fast and effective lookup, the key $\mathbf{K}$ is embedded into a dense vector using a text embedding model $\mathcal{E}$ and cached in the LTM. In this way, the LTM facilitates lifelong learning by transforming transient episodic memories into reusable, generalizable knowledge.

\subsection{PragmaBot Algorithm}
\label{sec:loop}

When presented with a new task, the task planner $\mathcal{P}$ leverages past experience through a retrieval-augmented generation (RAG) mechanism. A retrieval key $\mathbf{K}'$ is constructed from the current instruction $\mathbf{I}$ and the initial scene description. Using the text-embedding model $\mathcal{E}$, the system computes the embedding of $\mathbf{K}'$ and retrieves the top-$k$ most similar past experiences from the LTM $\mathbf{M}$ via cosine similarity:
\begin{equation}
    \{(\mathbf{K}_i, \mathbf{E}_i)\}_{i=0}^{k-1} \gets \operatorname*{arg\,top\,k} \limits_{(\mathbf{K}, \mathbf{E}) \in \mathbf{M}} \, \left[ \frac{ \mathcal{E}(\mathbf{K})^T\,\mathcal{E}(\mathbf{K}')} 
    {\|\mathcal{E}(\mathbf{K})\|\,\|\mathcal{E}(\mathbf{K}')\|} \right].
\end{equation}
These retrieved experiences are incorporated into the planning prompt, enabling $\mathcal{P}$ to leverage in-context learning by drawing on prior knowledge to make more informed decisions in new but related scenarios.

The STM, initialized as an empty set, is provided to the robot for online adaptation. At each time step, the task planner $\mathcal{P}$ (the high-level policy to be learned) generates an action based on the current observation, the current STM, and the retrieved LTM entries. After execution, the success detector $\mathcal{R}$ evaluates the action outcome and provides a success or failure signal as the feedback. Upon action failure, the planner performs self-reflection and takes a step along the linguistic gradient to select the next action. Upon task completion, the STM is summarized and stored in LTM for future use.

\begin{figure}[!t]
    \centering
    \includegraphics[width=\columnwidth]{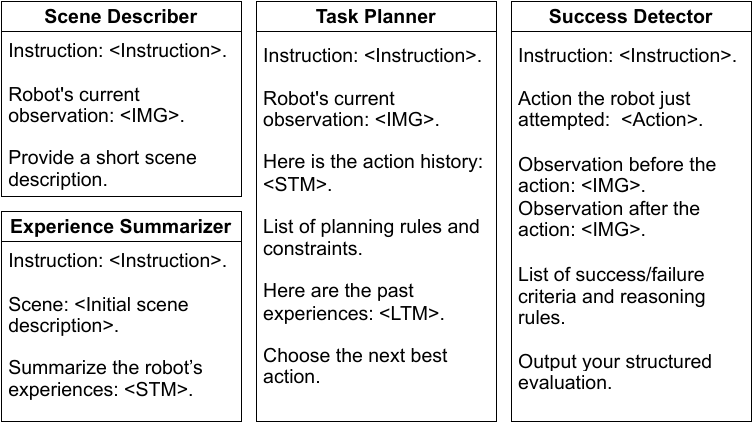}
    \caption{\textbf{Prompt templates used in different VLM modules.}}
    \label{fig:prompt}
    \vspace{-1.5em}
\end{figure}

This closed-loop process---combining retrieval from long-term memory, in-context adaptation via short-term memory, and experience accumulation---forms the core of the \shorttitle framework. It enables both rapid initialization on new tasks and continuous improvement over time, supporting effective lifelong learning in real-world environments. The complete algorithm is outlined in \Cref{alg:expteach}. The prompt templates for VLM modules are shown in \Cref{fig:prompt}; full examples are available on the project webpage.

\subsection{Enhanced Skillset with Image Annotations}
\label{sec:annotation}

For the low-level skills, we focus on three manipulation ones---{pick}, {place}, and {push}. Identifying a suitable location for each action is non-trivial: purely geometric grasp planning may seize an undesirable part (e.g., the meat on a skewer or the ice-cream top rather than its cone), and effective placing or pushing likewise demands semantic scene understanding. To address these challenges, we introduce an {on-demand image-annotation tool} that is shared across all skills.  Given a user instruction $\mathbf{I}$ and the current RGB frame $\mathbf{o}_t$, VLM first selects the skill to execute with its parameters (e.g., object name, whether image annotation is needed). 
The robot then performs open-vocabulary segmentation with {Grounded~SAM}~\cite{ren2024grounded}, which integrates Grounding~DINO~\cite{liu2024grounding} and SAM~\cite{kirillov2023segment}, producing an initial object mask. If the VLM requests a second annotation pass, our image overlays a set of candidate location masks on the image, allowing the VLM to choose the most appropriate location for the current action. 

\begin{figure}[!t]
    \centering
    \includegraphics[width=\columnwidth]{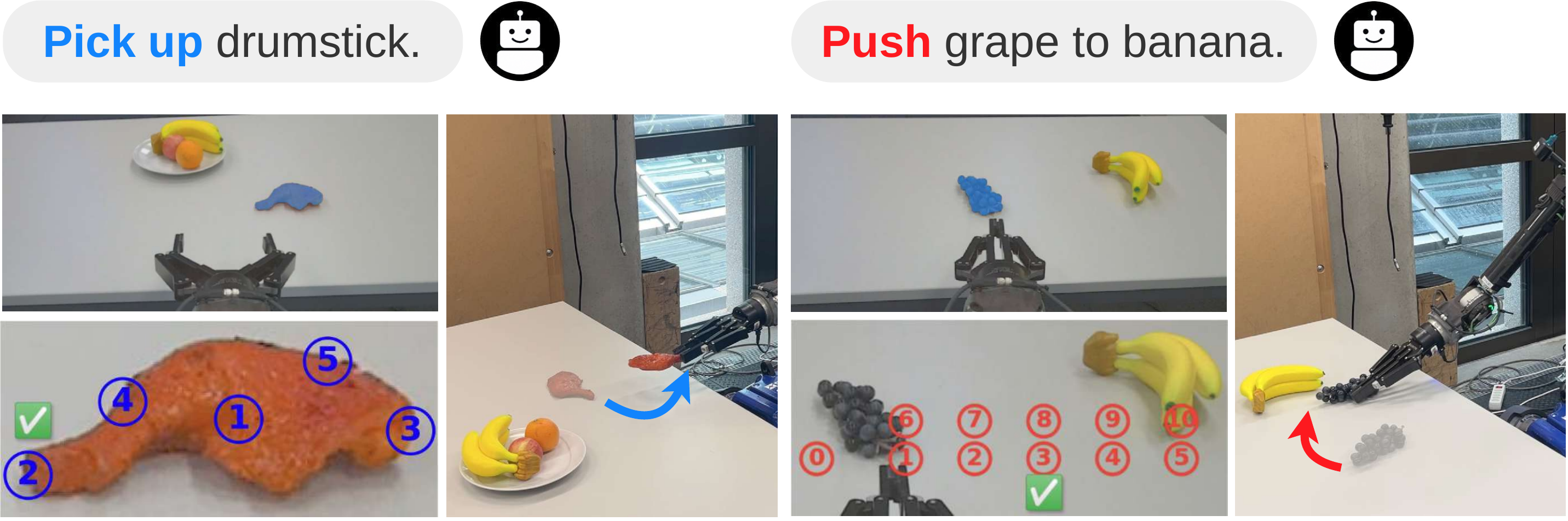}
    \caption{\textbf{Illustration of image annotation tools.}}
    \label{fig:pipeline_image_annotation}
\vspace{-1.5em}
\end{figure}

For placing, we apply farthest-point sampling (FPS)~\cite{eldar1997farthest} on the segmented mask to generate candidate placement locations, whereas for pushing, we draw candidate goal masks that denote the end points.
In both cases, the VLM evaluates the annotated options and selects the most suitable location, as illustrated in \Cref{fig:pipeline_image_annotation}. For grasping, the segmented point cloud is first passed to {AnyGrasp}~\cite{fang2023anygrasp}, which returns a set of grasp hypotheses accompanied by confidence scores $s_{conf} \in [0, 1]$. Grasp poses that violate kinematic constraints are filtered out through inverse-kinematics checks using {Pinocchio}~\cite{carpentier2019pinocchio}, yielding the feasible subset $\mathcal{G}$. If the VLM determines that image annotation is beneficial (typically not the case for simple objects like an apple), our annotation tool similarly performs FPS within the object mask to generate a collection of numbered location masks for the VLM to select from. The final grasp is selected by maximizing the product of two scores:
\begin{equation}
    g^* = \argmax\nolimits_{g \in \mathcal{G}} s_{conf}(g) \cdot s_{loc}(g),
\end{equation}
where $s_{loc}(g) \in [0, 1]$ denotes the location score, computed based on the normalized Euclidean distance between grasp $g$ and the chosen location.

\section{Results}
\subsection{Experiment Setup}

In our experiments, we use a legged manipulator that combines ANYmal~\cite{hutter2016anymal}, a quadrupedal robot, with a 6-DoF arm. The arm is equipped with a Robotiq 2F-140 gripper for object manipulation and a ZED X Mini Stereo Camera mounted on the elbow for perception. We employ \texttt{gpt-4o}~\cite{achiam2023gpt} for the VLM model and use OpenAI's \texttt{text-embedding-3-large}~\cite{openai2024embedding} for the text embedding model $\mathcal{E}$. If the robot fails an action and alters the environment, a human operator can choose to reset the scene, after which the robot resumes execution. 

\begin{figure*}[t]
\centering
\begin{subfigure}{0.24\textwidth}
\centering
\includegraphics[width=\linewidth]{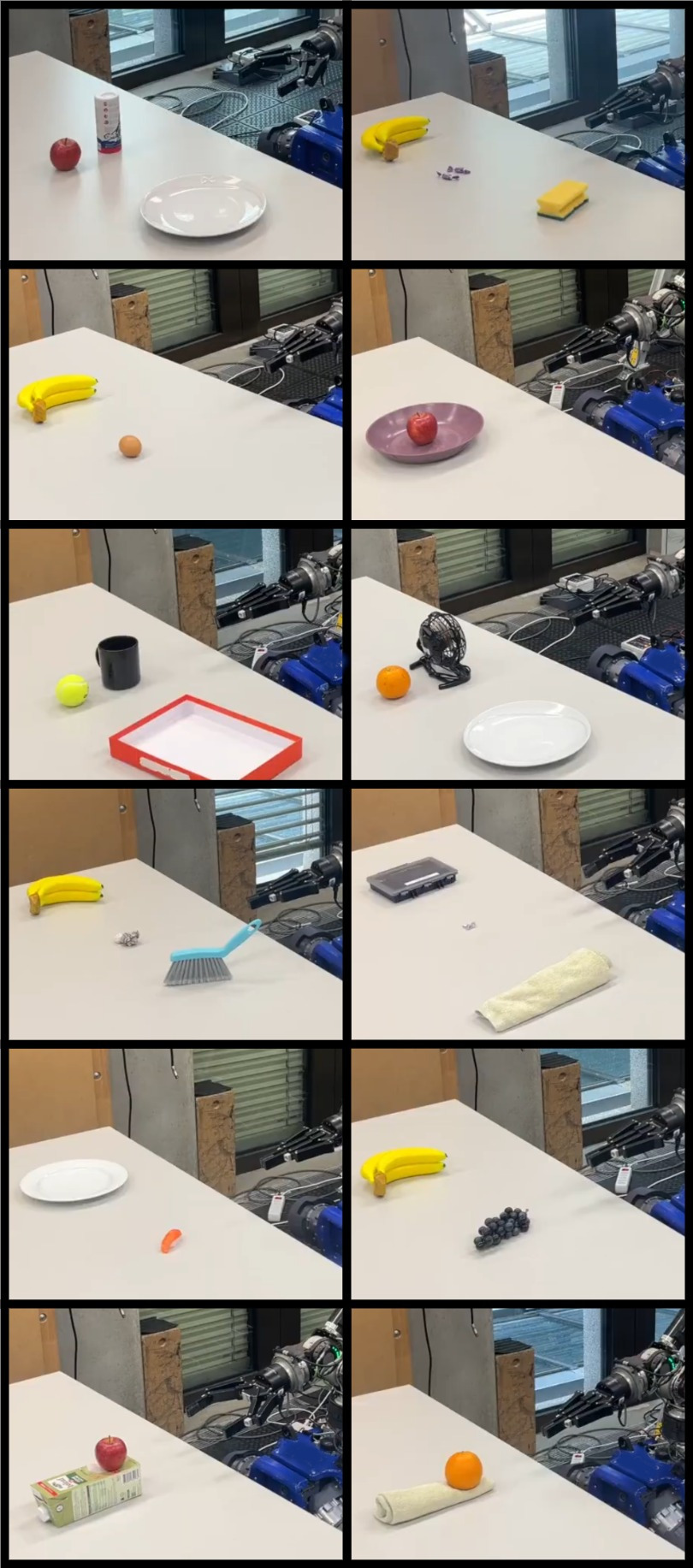} 
\caption{Illustration of all scenes.}
\label{fig:all-scenes}
\end{subfigure}
\hfill
\begin{subfigure}{0.75\textwidth}
\centering
\includegraphics[width=\linewidth]{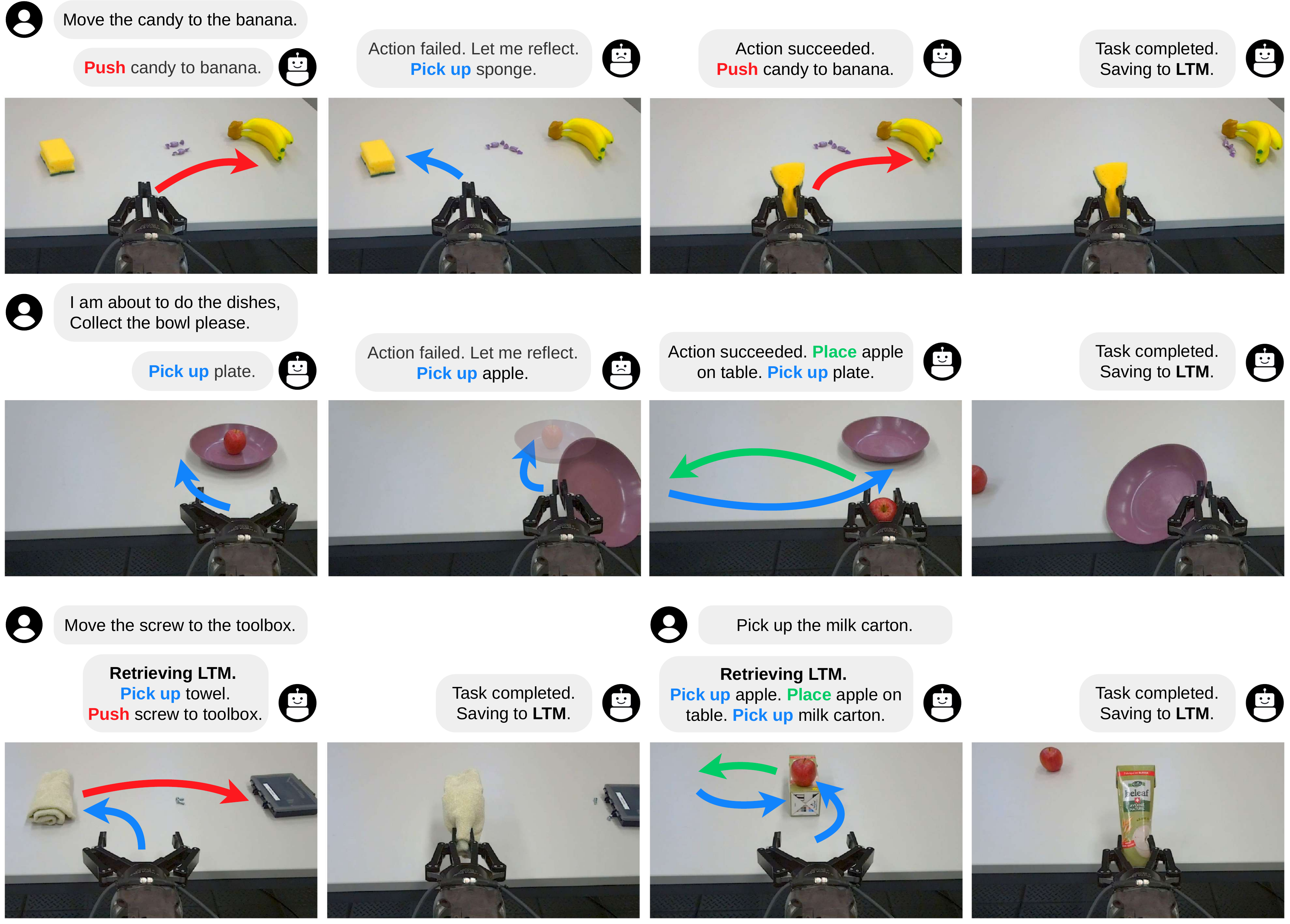}
\caption{Examples of \shorttitle on different scenarios.}
\label{fig:stm-ltm-examples}
\end{subfigure}
\caption{\textbf{Overview of all experimental scenes (a) and demonstration of \shorttitle's performance across four representative scenarios (b).} In the first two examples (top and middle rows), the robot successfully completes the tasks after self-reflection. These experiences are then summarized and stored in LTM, enabling the robot to generalize its learning to similar future scenarios (bottom row).}
\label{fig:stm_ltm_results}
\vspace{-1.5em}
\end{figure*}

\subsection{Evaluation of Short-Term Memory and Self-Reflection}
\label{sec:stm_evaluation}

To evaluate the effectiveness of the STM and reflection module at efficiently generating successful episodes, even after initial failures, we designed four challenging object manipulation tasks (top two rows in~\Cref{fig:all-scenes}). For the baseline, we use CaP-V, which enhances CaP~\cite{liang2023code} by incorporating visual feedback. 
Note that CaP-V
does not have STM and selects the next action solely based on the current image and user instruction, without the ability to reflect on failed actions. Experimental results in~\Cref{tab:stm_result} highlight the critical role of STM and reflection in achieving successful task completion. Without STM, the robot tends to repeat the same failures without adapting, leading to poor task performance.

\begin{table}[!t]
    \centering
    \caption{\textbf{Effect of STM on task success rates.} Each task is tested 5–10 times with two attempts allowed.}
    \begin{tabular}{lcc}
    \toprule
    \textbf{Task} & \textbf{CaP-V} & \textbf{\shorttitle} \\
    \midrule
    {Put apple on plate (container obstructs)} & 43\% & \textbf{86\%} \\
    \midrule
    {Move tiny candy (sponge/towel nearby)} & 22\% & \textbf{67\%} \\
    \midrule
    {Move egg (open view)} & 40\% & \textbf{100\%} \\
    \midrule
    {Pick up bowl (apple inside)} & 33\% & \textbf{83\%} \\
    \bottomrule
    \end{tabular}
    \label{tab:stm_result}
    \vspace{-1.5em}
\end{table}

With STM, the robot can successfully reflect on its failures, leading to emergent intelligent object interactions (including tool use) and ultimately task success. For example, when instructed to ``put the apple on the plate,'' and faced with a partially obstructing container, the robot initially fails to grasp the apple due to poor perception and obstruction. After detecting the failure, the VLM decides to push away the container and successfully retries the grasp (see~\Cref{fig:teaser}).
When asked to ``collect the bowl,'' where an apple is already inside, the robot initially tries to lift the bowl directly. However, the apple falls out during the action execution, and the robot reflects and revises its plan: ``I should first move the apple to the table before picking up the bowl'' (see~\Cref{fig:stm-ltm-examples}).
Similarly, when told to ``move the candy to the banana,'' the robot fails to push the candy with its gripper due to insufficient contact. Upon reflection, the VLM autonomously chooses to use a sponge as a tool to push more effectively (see~\Cref{fig:stm-ltm-examples}). 
After cracking an egg while grasping, the VLM similarly learns to push rather than grasp fragile objects.

Note that these instances of autonomous intelligent adaptation would not be possible without the high accuracy of the VLM’s success detection. In object-picking tasks, we observe a false negative rate of 6.67\% (4/60) and a false positive rate of 5\% (3/60), with most false positives occurring when the object remains on the table but appears visually enclosed in the gripper.

\subsection{Evaluation of Long-Term Memory and Generalization}
\label{sec:ltm_evaluation}

Upon completing the task in~\Cref{sec:stm_evaluation}, the STM is autonomously transferred to the LTM for future use. Together with 96 limited instructional experiences from simpler tasks, we build an LTM containing 100 entries. For the baseline, we use COME~\cite{zhi2024closed}, which does not have access to the LTM. We keep this generated LTM frozen during evaluation for a fair comparison. We first evaluate the effectiveness of the LTM in the same four scenes with the same objects. As shown in~\Cref{tab:ltm_result}, the robot successfully recalls the correct action from the LTM with RAG, demonstrating improved performance compared to COME. We further investigate its ability to generalize to other similar scenarios. To do so, we modify the scene to create new but structurally similar scenarios, as illustrated in~\Cref{fig:all-scenes} (bottom four rows). The results in~\Cref{tab:ltm_result} show that experience gained from one task successfully transfers to related tasks under similar conditions and significantly improves the task success rate. For example, when asked to ``Move the screw to the toolbox,'' the robot immediately decides to use a towel to push the screw successfully. Similarly, when tasked with ``Pick up the milk carton,'' it remembers to reposition the apple first before retrieving the target item.

\begin{table}[!t]
    \centering
    \caption{\textbf{Effect of LTM on single-trial task success rates.} Each task is tested 5–10 times.}
    \begin{tabular}{lcc}
        \toprule
        \textbf{Task} & 
        \textbf{COME~\cite{zhi2024closed}} & 
        \textbf{\shorttitle} \\
        \midrule
        Put apple on plate ({container obstructs}) & 29\% & \textbf{100\%} \\
        Move tiny candy ({towel nearby})                            & 11\% & \textbf{78\%} \\
        Move egg ({open view})                             & 20\% & \textbf{100\%} \\
        Pick up bowl ({apple inside})     & 17\% & \textbf{83\%} \\
        \hline
        Put tennis ball in box ({mug obstructs})           & 29\% & \textbf{71\%} \\
        Put orange/ball on plate ({fan blocks})    & 10\% & \textbf{80\%} \\
        Move crumpled paper ({brush nearby}) & 25\% & \textbf{63\%} \\
        Move screw ({towel nearby})  & 0\% & \textbf{86\%} \\
        Move sushi ({open view})                         & 14\% & \textbf{71\%} \\
        Move grape/cherry ({open view})                    & 20\% & \textbf{70\%} \\
        Pick up box ({apple on top})      & 43\% & \textbf{86\%}  \\
        Pick up towel ({orange on top})   & 50\% & \textbf{75\%}  \\
        \bottomrule
    \end{tabular}
    \label{tab:ltm_result}
    \vspace{-2em}
\end{table}

\begin{figure}[!t]
    \centering
    \includegraphics[width=\columnwidth]{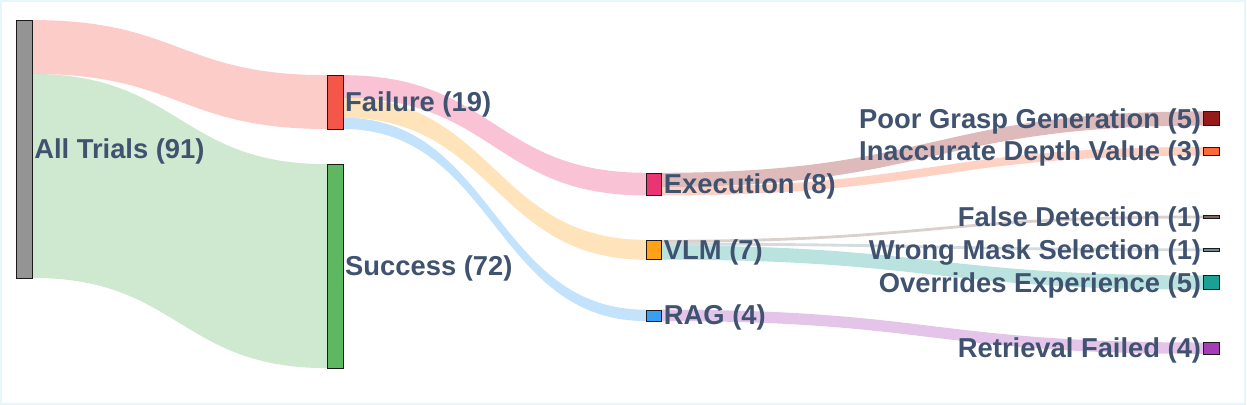}
    \caption{\textbf{Failure flow diagram illustrating the sources of initial failure across two hierarchical levels.}}
    \label{fig:failure-flow-diagram}
    \vspace{-1.5em}
\end{figure}

We analyzed the 19 first-failure cases out of 91 total trials, as shown in~\Cref{fig:failure-flow-diagram}. Of these failures, 8 stem from action execution issues---either poor grasp generation or inaccurate depth estimates from the stereo camera. Another 7 failures arise from VLM reasoning errors, most notably when the VLM perceives the retrieved memory as misaligned with the current visual observation and consequently downweights or ignores the relevant experience. For instance, in the ``Put tennis ball in the box (mug obstructs)'' scenario, the retrieved memory involved clearing a similar occlusion, but the VLM might deem the mug’s obstruction negligible and overrode that strategy, leading to failure. The remaining 4 failures occur when the RAG fails to retrieve the relevant experience.

\subsection{Ablation Study of Memory Retrieval}

\begin{figure}[!t]
    \centering
    \includegraphics[width=\columnwidth]{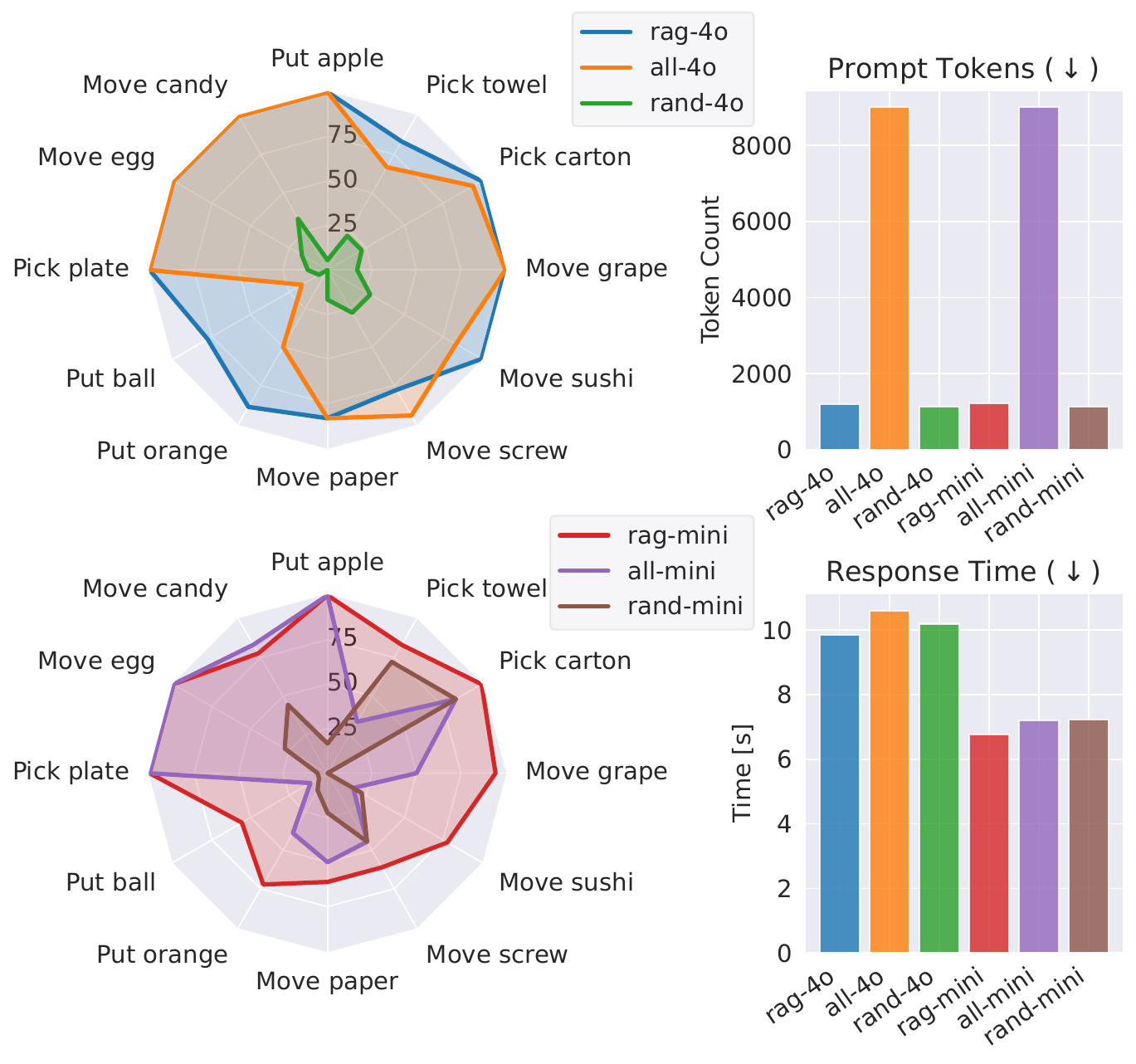}
    \caption{\textbf{Ablation study of the memory retrieval module.} The radial axes of the radar charts represent the accuracy of the first planned action in percentage. RAG with \texttt{gpt-4o} (rag-4o) provides the highest accuracy.}
    \label{fig:rag-ablation-study}
    \vspace{-1em}
\end{figure}

To evaluate the effectiveness of our memory retrieval strategy, we measure the accuracy of the first planned action (without execution) across 12 tasks under three retrieval settings, as shown in~\Cref{fig:rag-ablation-study}. We reuse the LTM described in~\Cref{sec:ltm_evaluation}. Randomly selecting $k=5$ memories yields the worst performance, with an average accuracy of only $17\%$ on unseen tasks, as task-relevant experiences are rarely retrieved by chance. Providing the entire LTM improves accuracy to $74\%$, but unfiltered retrieval introduces irrelevant or distracting information, leading to unstable behavior. This aligns with prior findings that excessively long or noisy contexts can degrade model performance, as LLMs may struggle to focus on the most relevant content~\cite{liu2024lost,shuster2021retrieval}. In contrast, our RAG-based retrieval strategy achieves the highest accuracy at $89\%$. We also evaluated \shorttitle using the smaller model \texttt{gpt-4o-mini}. We found that this smaller model tends to behave more conservatively, making it more aligned with our robot when it does not have access to relevant experiences. When provided with relevant memories, \texttt{gpt-4o-mini} also shows clear performance gains, though the improvement is less pronounced than for larger models. Consequently, both {rag-4o} and {all-4o} outperform their mini counterparts, albeit at the cost of higher latency. Moreover, feeding the full LTM increases text prompt length by a factor of $7.5$ (excluding images), resulting in substantially higher monetary cost.

\subsection{Ablation Study of Image Annotation Module}

\begin{figure}[!t]
    \centering
    \includegraphics[width=\columnwidth]{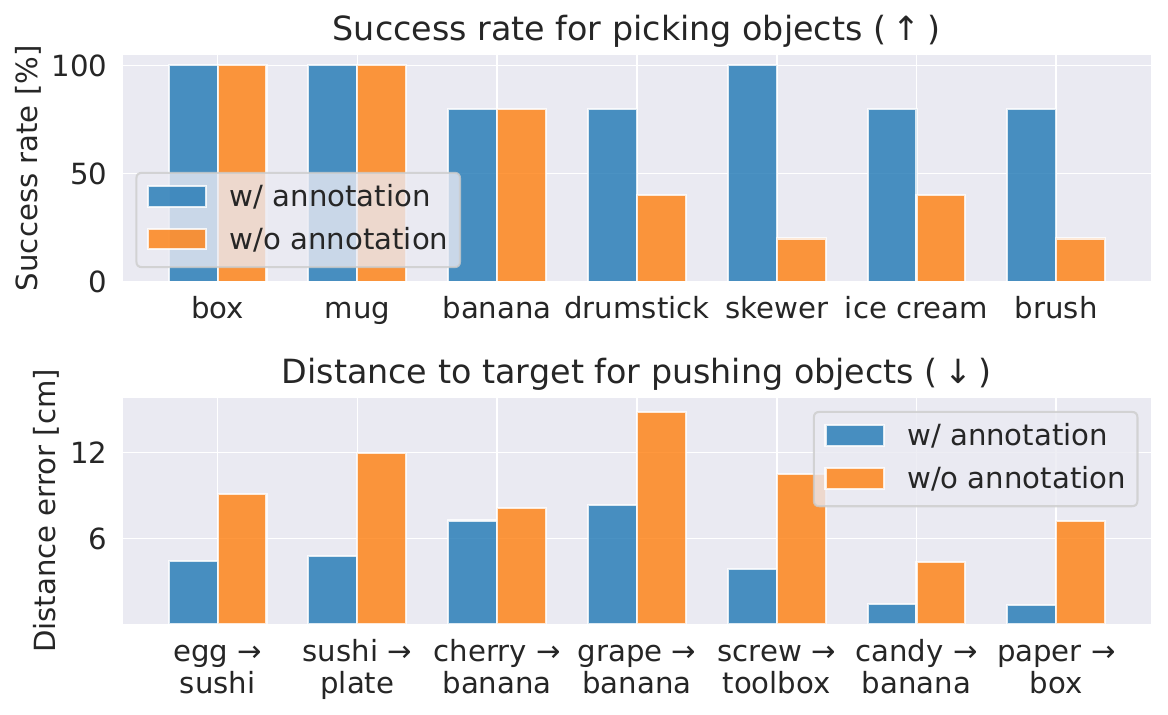}
    \caption{\textbf{Ablation study of the image annotation module.} \textit{Top:} success rates for picking objects (higher is better).
    \textit{Bottom:} distance errors in pushing one object to another (lower is better). The results demonstrate that incorporating image annotation consistently enhances performance.}
    \label{fig:result_image_annotation}
    \vspace{-1em}
\end{figure}

To evaluate the effectiveness of image annotation on grasping tasks, we compared success rates with and without it across 7 objects. A grasp was considered successful if it targeted the correct object section (e.g., the stick of a meat skewer). As shown in \Cref{fig:result_image_annotation}, annotation significantly improves success rates for objects with complex shapes that require grasping specific sections (e.g., drumsticks, skewers). Without annotation, AnyGrasp~\cite{fang2023anygrasp} often favors larger surfaces due to its reliance on geometric cues. The failures with annotation are mostly due to inaccurate 3D point clouds. We also evaluated pushing by measuring the distance error to the target location and found that image annotation consistently reduces this error, highlighting its benefits beyond grasping.
While effective, this method introduces additional latency, requiring an average of \SI{5.15}{\second} to retrieve a response from \texttt{gpt-4o}.

\section{Conclusion and Future Work}

This work introduces \shorttitle, a novel method that enables robots to learn to plan tasks by experiencing the real world. Empirical results demonstrate that \shorttitle allows robots to autonomously reflect and adapt using short-term memory, significantly improving task success rates and facilitating intelligent object interactions---such as creative tool use. Furthermore, these experiences can be stored in the long-term memory, enabling the robot to plan correctly on its first attempt in the future, even in previously unseen scenarios. Extensive evaluations show that integrating this persistent memory with RAG yields substantial performance gains over current state-of-the-art methods across 12 challenging scenarios. Our framework provides a general and efficient paradigm that does not rely on substantial training resources, making it well-suited for deployment in real-world robotic systems.

While \shorttitle demonstrates strong potential in learning task planning, it still presents several limitations. In practice, certain real-world scenarios involve information that cannot be captured by vision alone. Integrating additional modalities--such as tactile or auditory signals--could greatly improve the system's capacity to interpret complex, multimodal feedback. Second, our current memory database is limited by the cost of real-world data collection. Scaling to much larger memory databases raises important questions: should memory pruning be applied when the database becomes extremely large? If so, which memories should be retained and which forgotten? Could the VLM itself perform filtering without relying heavily on human-designed heuristics? Moreover, as the memory grows, our current top-$k$ retrieval mechanism may become ineffective, warranting exploration of more sophisticated strategies---such as maximum marginal relevance (MMR). Finally, sharing memories among multiple robots is a promising direction. This is clearly feasible when robots share identical hardware and skillsets, but what if  they have similar morphologies (e.g., another quadrupedal robot with a single arm, but much smaller)? Could a robot initially attempt to apply transferred memories and, upon failure, discard unsuitable experiences and replace them with its own? These are compelling directions for future work.

\bibliographystyle{IEEEtran}
\bibliography{reference}


\end{document}